\documentclass[sigconf]{aamas} 

\usepackage{balance} 

\usepackage{hyperref}       
\usepackage{url}            
\usepackage{amsfonts}       
\usepackage{nicefrac}       
\usepackage{microtype}      

\usepackage{comment}
\usepackage{lipsum}
\usepackage{amsmath}
\usepackage{amsthm}
\usepackage{bm}
\usepackage{color}
\usepackage{multirow}
\usepackage{here}
\usepackage{balance}

\newcommand{\x}{\bm{x}}
\newcommand{\X}{\bm{X}}
\newcommand{\z}{\bm{z}}
\newcommand{\Z}{\bm{Z}}
\newcommand{\yf}{y^{\rm f}}
\newcommand{\ycf}{y^{\rm cf}}
\newcommand{\bk}{\bm{k}}

\newcommand{\ba}{\bm{a}}

\theoremstyle{plain}
\newtheorem{assumption}{Assumption}

\setcopyright{ifaamas}
\acmConference[AAMAS '21]{Proc.\@ of the 20th International Conference on Autonomous Agents and Multiagent Systems (AAMAS 2021)}{May 3--7, 2021}{Online}{U.~Endriss, A.~Now\'{e}, F.~Dignum, A.~Lomuscio (eds.)}
\copyrightyear{2021}
\acmYear{2021}
\acmDOI{}
\acmPrice{}
\acmISBN{}

\acmSubmissionID{fp221}

\title{Grab the Reins of Crowds: Estimating the Effects of Crowd Movement Guidance Using Causal Inference}

\author{Koh Takeuchi}
\affiliation{
  \institution{Kyoto University}
  }
\email{takeuchi@i.kyoto-u.ac.jp}

\author{Ryo Nishida}
\affiliation{
  \institution{Tohoku University, AIST}
  }
\email{ryo.nishida@aist.go.jp}

\author{Hisashi Kashima}
\affiliation{
  \institution{Kyoto University}
  }
\email{kashima@i.kyoto-u.ac.jp}

\author{Masaki Onishi}
\affiliation{
  \institution{AIST}
  }
\email{onishi@ni.aist.go.jp}

\begin{abstract}
Crowd movement guidance has been a fascinating problem in various fields, such as easing traffic congestion in unusual events and evacuating people from an emergency-affected area.
To grab the reins of crowds, there has been considerable demand for a decision support system that can answer a typical question: {\em ``what will be the outcomes of each of the possible options in the current situation?"}. 
In this paper, we consider the problem of estimating the effects of crowd movement guidance from past data.
To cope with limited amount of available data biased by past decision-makers, we leverage two recent techniques in deep representation learning for spatial data analysis and causal inference.
We use a spatial convolutional operator to extract effective spatial features of crowds from a small amount of data and use balanced representation learning based on the integral probability metrics to mitigate the selection bias and missing counterfactual outcomes.
To evaluate the performance on estimating the treatment effects of possible guidance, we use a multi-agent simulator to generate realistic data on evacuation scenarios in a crowded theater, since there are no available datasets recording outcomes of all possible crowd movement guidance.
The results of three experiments demonstrate that our proposed method reduces the estimation error by at most $56\%$ from state-of-the-art methods.
\end{abstract}


\newcommand{\BibTeX}{\rm B\kern-.05em{\sc i\kern-.025em b}\kern-.08em\TeX}

\begin{document}


\pagestyle{fancy}
\fancyhead{}


\maketitle 


\begin{figure}[t]
  \centering
  \includegraphics[width=8.5cm]{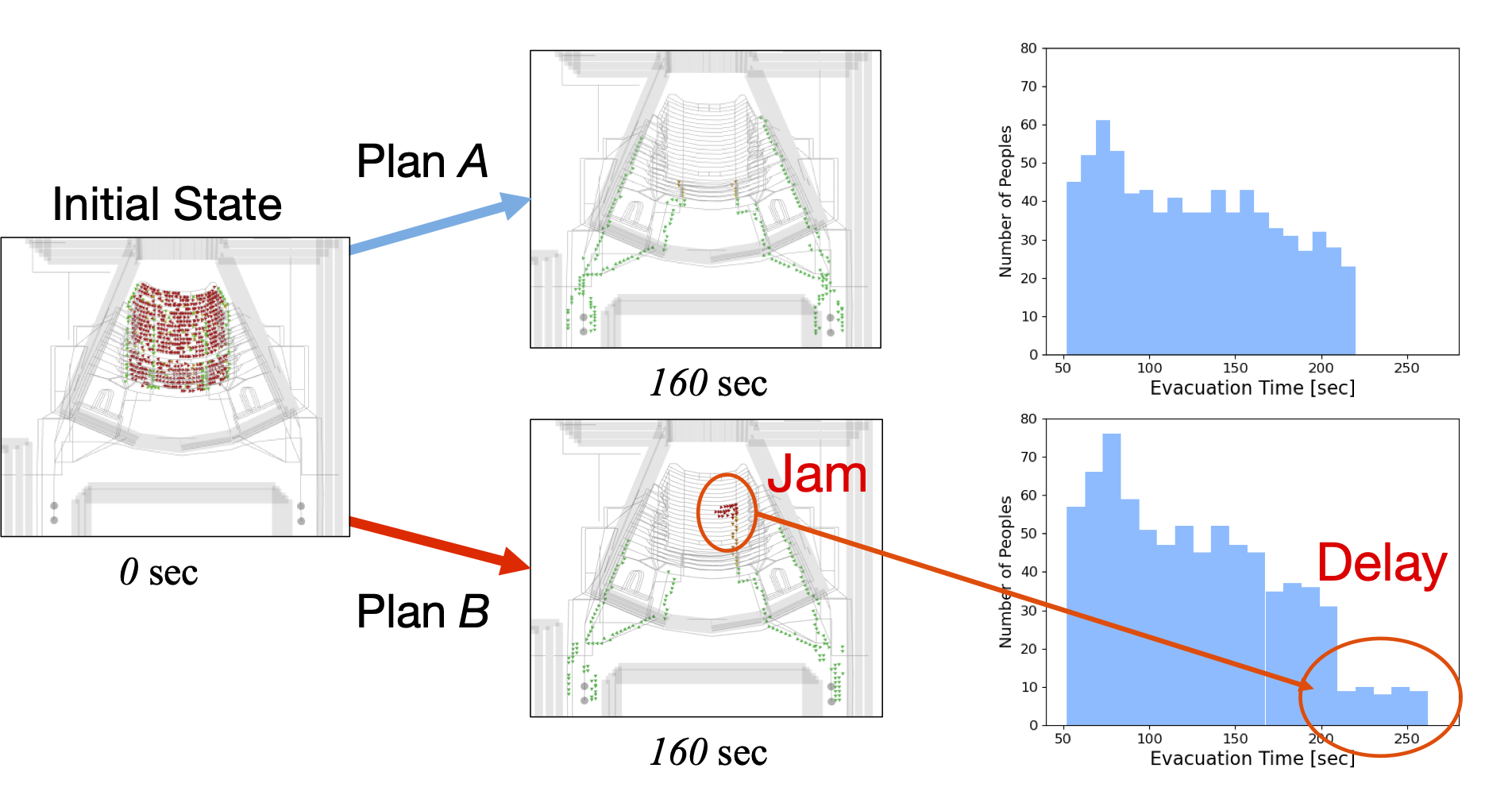}
  \caption{Example of crowd movement guidance: Evacuation events from an opera house. The green and red dots represent people who are moving or stopping, respectively.}
  \label{fig:treatmenteffect}
  \Description{guidance}
\end{figure}

\section{Introduction}
Crowd movement guidance has been a fascinating problem in various fields, such as easing traffic congestion in normal or abnormal events \cite{varakantham2015direct,sharon2017real,singh2019multiagent} and evacuating people from an emergency-affected areas such as a crowded building \cite{lammel2008bottlenecks,gianni2008simulation,tsai2011escapes,8823571,yadav2016using}.
To grab the reins of crowds,  given a situation, decision-makers need to devote extensive efforts to quickly select guidance that consists of a complex compound of actions.
For example, a traffic network manager needs to control traffic signals, road capacities, and directions of crowd flow together while considering the interactions among them.
We show an example of crowd movement guidance in Figure \ref{fig:treatmenteffect}.
Let us assume a situation in which theater managers need to decide guidance required for a crowd evacuation, and there are two guide plans: {\em A} and {\em B}.
If they choose plan {\em A}, congestion of crowds does not occur, and thus people can quickly escape from the theater.
In contrast, plan {\em B} disturbs the crowd movements, which results in longer evacuation times than in plan {\em A}.
The choice of plan has a significant impact on the outcome, and therefore, there has been considerable demand for a decision support system that can answer a typical question posed by decision-makers: {\em ``what will be the outcomes of each of the possible options in the current situation?"}. 

However, due to the limited number of such events or evacuation situations, there are not enough data records for estimating such effects with sufficient accuracy.
To make matters worse, these records contain biases introduced by the plans taken by past decision-makers, because they try to make the best decisions for the situations at the time, and we cannot expect randomized control trials (RCTs) were performed to obtain unbiased data due to ethical or other reasons.
In addition, due to the counterfactual nature of the data acquisition process, we cannot observe guidance outcomes other than those of the actual ones, and therefore, we cannot directly assess the effects of guidance from such datasets.

In this paper, we focus on the problem of estimating the effects of crowd movement guidance from a relatively small amount of biased data.
We propose a {\em spatial convolutional counterfactual regression} (SC-CFR), which takes advantage of two recent developments in deep representation learning in spatial data analysis and causal inference.
Because the locational distribution of crowds has a significant impact on the effectiveness of movement guidance, 
SC-CFR aggregates crowd distribution information by means of the spatial convolutional layer \cite{ma2017learning,10.5555/3298239.3298479} and predicts the outcomes of movement guidance with high accuracy even with a limited number of data.
On the other hand, to cancel the data biases made by past decision-makers, we resort to the causal inference framework that has been successfully applied to answer the what-if problems in various domains, such as healthcare \cite{shalit17a}, economics \cite{Imbens2009}, and education \cite{zhao2017estimating}.
In particular, SC-CFR employs balanced representation learning based on integral probability metrics, which mitigates selection bias and missing counterfactual outcomes \cite{shalit17a}.

Evaluating the performance of estimating the possible guidance's treatment effects requires datasets recording outcomes of all possible crowd movement guidance. 
Since there are no datasets available for our purpose, we used a multi-agent simulator \cite{Yamashita2012CrowdWalk} to generate realistic data on evacuation scenarios in a crowded theater.
From these datasets, we conducted experiments for estimating the treatment effects on outcomes, such as the maximum, average, and standard deviation of evacuation times required by crowds.
We demonstrate that our proposed method reduced the estimation error by at most $56\%$ from state-of-the-art baseline methods that do not consider spatial attributes.

Our contributions are summarized as follows:
\begin{itemize}
    \item We first address the what-if problem in crowd movement guidance as causal effect estimation from a relatively small amount of biased data.
    \item We propose SC-CFR, which takes advantage of two recent developments in deep representation learning in spatial data analysis and causal inference.
    \item We demonstrate that our method reduces the estimation error by at most $56\%$ from state-of-the-art baselines by experiments with a multi-agent simulator.
\end{itemize}


\begin{figure}[tb]
  \centering
  \includegraphics[width=7cm, trim=0mm 10mm 0mm 0mm]{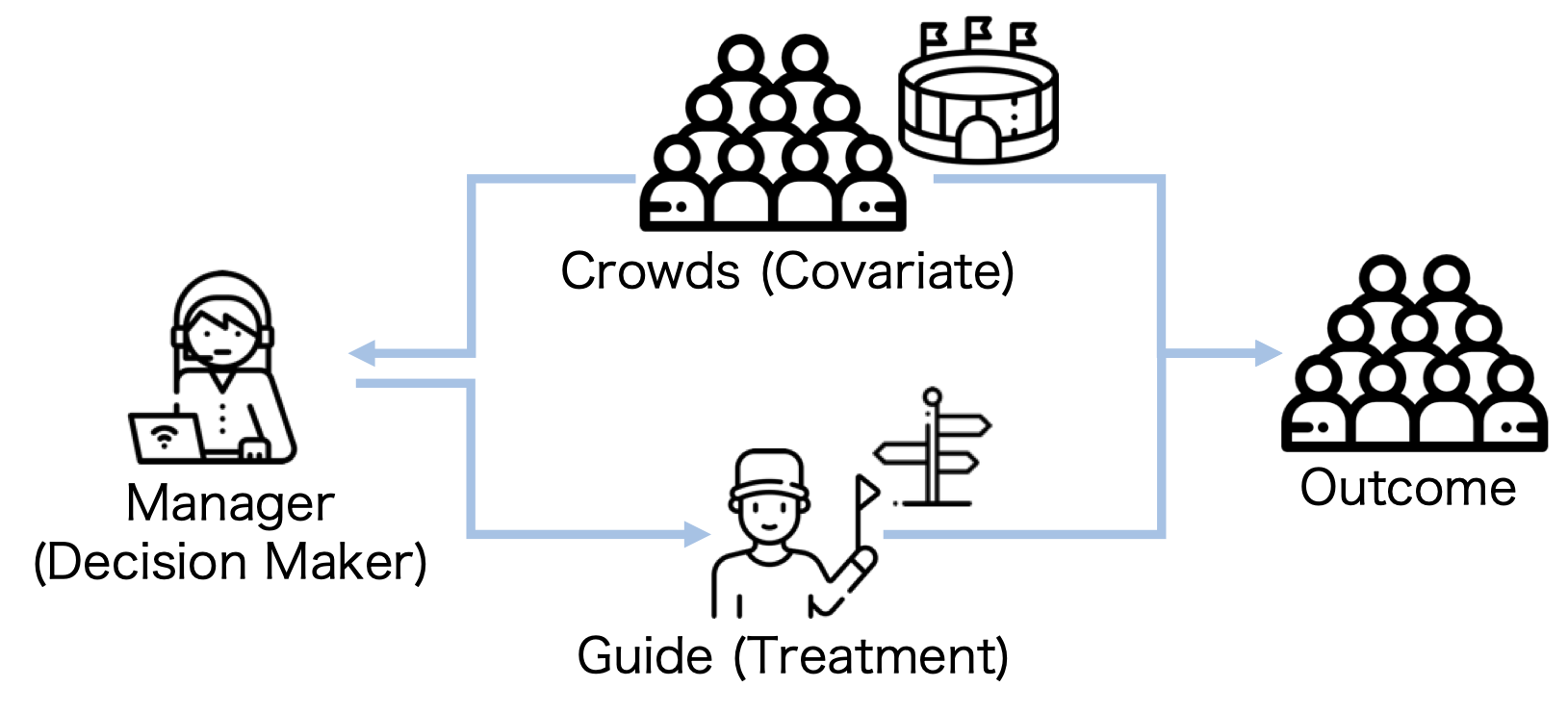}
  \caption{Framework of crowd movement guidance based on causal inferences.}
  \label{fig:guidance}
  \Description{guidance}
\end{figure}

\section{Preliminaries}
We use an uppercase letter to denote a random variable (i.e., $X$), a lowercase letter to indicate a realization $x$ drawn from a particular distribution (i.e., $x \sim P(X)$ ), and a calligraphy letter $\mathcal{X}$ to represent a sample space (i.e., $x \in \mathcal{X}$ ). 

\subsection{Crowd Movement Guidance}
Let us consider a place or area with crowds of freely moving people.
When the place is crowded, a manager should want to carry out a crowd guidance plan to reduce the congestion caused by crowds moving toward other places.
For example, if a park hosts a big festival with a sell-out crowd, a park manager has to make a decision on guidance to avoid traffic jams.
Another example is of a manager working in a stadium who needs to make a decision on guidance to safely and quickly evacuate crowds during a fire emergency.
A crowd guidance plan can comprise various types of actions, such as route recommendations for crowds or a modification of road width, which could have complicated interactions. 

Assume that a situation of crowds, such as their current locations, is observed as covariates.
We denote a situation of crowds (covariates) as $\X \in \mathbb{R}^d$, where $d$ is the number of people.
A guide (treatment) is $\Z \in \mathcal{Z} \subset \{0,1\}^t $ that consists of $t$ possible actions, where $\mathcal{Z}$ is a set of possible guides.
A guide can select single or multiple actions (i.e., a combination of actions); therefore, $|\mathcal{Z}|$ grows exponentially as $t$ increases.
We denote a potential outcome as $Y \in \mathbb{R}$, such as an average moving cost or the maximum movement cost consumed by crowds.
For example, let us suppose an evacuation scenario from a place.
Given an initial situation of crowds $\x \sim P(\X)$, a decision-maker, who needs to guide crowds to a safe place, selects a treatment $\z \sim P(\Z|\X)$, after which crowds start to move.
We can observe an outcome $y \sim P(Y|\X,\Z)$, such as evacuation times, after finishing the scenario.
This procedure is built upon by the Rubin-Neyman potential outcomes framework \cite{rubin2005causal}. 
We show a framework for crowd movement guidance in Figure \ref{fig:guidance}.

\subsection{Causal Effect Estimation}
Suppose that there is a healthcare record (covariate) of a patient, and a doctor needs to choose whether to prescribe medicine, which we will call a treatment. 
We can observe a future health condition of the patient as a subsequent outcome.
Causal effect estimation is a task for estimating outcomes given not actual treatments and to assess the difference in outcomes between treatments.
In this paper, we condider the situation of crowds as covariates, guidance as a treatment, and costs of crowd movements as outcomes.
Effects of a guide should differ from situation to situation since it depends on the crowd conditions and types of events.
Thus, decision-making on guide planning is one of the most difficult tasks for managers, and they need a useful tool to help their decision-making.
Surprisingly, the effect of guidance has not been considered in the existing literature.

We suppose that two arbitrary guide plans {\em A} and {\em B} correspond to guides $\z$ and $\z'$, respectively.
Given covariates $\x$ and a guide $\z$, we denote the cost consumed by crowds as $Y_{\x}(\z)$.
The conditional average treatment effect (CATE) for covariates $\x$ can be defined as $\mathbb{E}[Y_{\x}(\z)] - \mathbb{E}[Y_{\x}(\z')]$, where we use $\mathbb{E}$ to denote the expectation over $Y$. 
Intuitively, this quantity indicates how much of an advantage (or disadvantage) plan {\em A} has in guiding the crowd movements compared with plan {\em B}.
Knowing this quantity helps a decision-maker compare the difference between treatments for a particular situation $\x$; however, we cannot directly calculate CATE from observed outcomes because we can only observe the outcome of one of the treatments.

\subsection{Selection Bias and Missing Counterfactuals}
Estimating the outcome of guidance requires a deep understanding of the complex relationships within covariates and guidance.
One of the difficulties in estimating treatment effects on crowd movement is that an observed set of guidance is biased in the given dataset, because the manager usually selects guidance according to their policy.
Thus, even if the observed covariates are random, we cannot observe the outcome with a pair of covariates and guidance that managers do not prefer.
Another obstacle is that we can only observe a factual outcome, but cannot observe counterfactual outcomes.
Usually, we can only deploy a single guide in a situation, and thus the outcomes for the other possible guides remain missing.

Formally, in an observed study, we can obtain only one factual outcome $\yf$ depending on a selected treatment $\z$, and thus never see counterfactual outcomes $\ycf$ corresponding to the other treatments that are not selected by the decision-maker. 
More formally, when a decision-maker chooses a treatment $\z$, a factual outcome $\yf$ is equal to the outcome $y_{\x}(\z)$, and the counterfactual outcomes $\ycf$ are $y_{\x}(\z')$, $\forall \z' \in \mathcal{Z} \setminus \z$.
We suppose that a decision-maker has a policy for choosing a treatment $\z$ based on a situation $\x$, and thus, observed treatments are not at random and always biased.
This bias, which is called a selection bias in the Rubin-Neyman framework, leads to a simple model for estimating outcomes, not accurate and biased results.
Thus, we need to develop a method that learns a model for predicting outcomes and can simultaneously cancel the effects of the selection bias.

In this paper, we make the common assumption called {\em strong ignorability} in the Rubin-Neyman framework.
\begin{assumption}[Stable Unit Treatment Value Assumption]\label{as:sutva} 
The potential outcomes for any situations do not vary with the treatment assigned to other situations, and, for each situation, there are no different forms or versions of each treatment level, which lead to different potential outcomes.
\end{assumption}
This assumption emphasizes the independence of each situation and that there are no interactions between situations. 
\begin{assumption}[Ignorability]\label{as:igno}
Given the covariate, $\X$, treatment assignment $\Z$ is independent to the potential outcomes, i.e., $\Z \perp Y_{\X}(\z) \mid \X, ~\forall z \in \mathcal{Z}$.
\end{assumption}
The ignorability assumption is also referred to the unconfoundedness assumption.
With this unconfoundedness assumption, the situations with the same covariates $\X$, their treatment assignment can be viewed as random.
\begin{assumption}[Positivity]\label{as:positivity}
For any value of $\X$, treatment assignment is not deterministic: $P(\Z=\z|\X=\x)>0, \forall\z$ and $\x$.
\end{assumption}
This assumption implies that all the factors determining the outcome of each treatment are observed.
From the above assumptions, we follow strong ignorability assumption that is referred to the no-hidden confounders assumption.
We formalize this assumption by using the standard strong ignorability condition: 
\begin{assumption}[Strong Ignorability]\label{as:strongigno}
 $\Z \perp Y_{\X}(\z) \mid \X, ~\forall z \in [0,1]^t$ and  $P(\Z=\z|\X=\x)>0, \forall\z$ and $\x$.
\end{assumption}
The strong ignorability assumption is a sufficient condition for CATE function $f$ to be identifiable \cite{Imbens2009}.


\begin{figure}[tb]
  \centering
  \includegraphics[width=7.5cm, trim= 20mm 5mm 30mm 5mm]{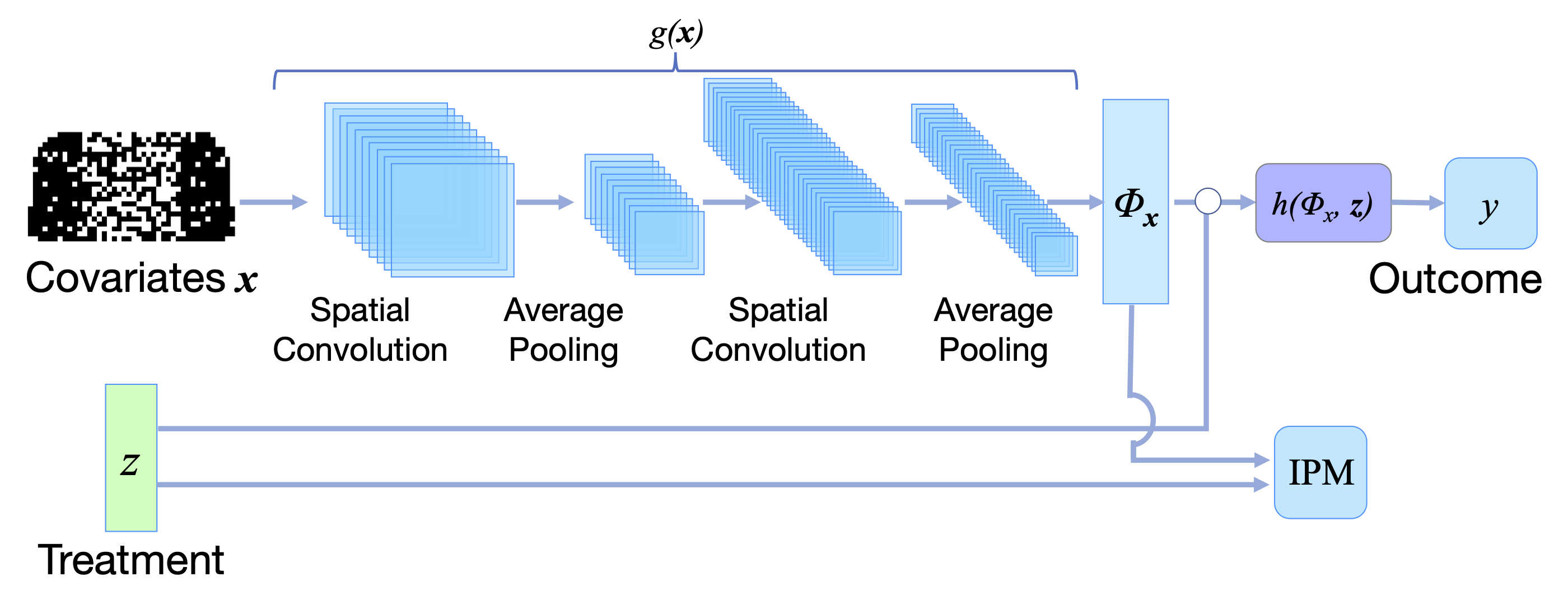}
  \caption{Model architecture of spatial convolutional counterfactual regression (SC-CFR).}
  \label{fig:model}
  \Description{our theater}
\end{figure}


\section{Spatial Convolutional Counterfactual Regression}
To predict what the outcomes of each of the possible options in the current situation will be, we develop a model $y=f(\x, \z)$ that predicts the outcome for any treatments $\z$.
Our idea is twofold:
a spatial convolutional operator extracting effective spatial features of crowds from a small amount of data, and balanced representation learning to mitigate the selection bias and missing counterfactuals. 

We assume that the biased observed dataset $\mathcal{D} = \{\x_n, \z_n, \yf_n\}_{n=1}^{N}$ is available for learning our model, where $N$ is the number of situations.
Note that, our dataset contains only factual outcomes for assigned treatments; hus, we do not have access to the unobserved counterfactual outcomes $\ycf$ for unassigned treatments.

\subsection{Model Architecture}
We first give an overview of our proposed model called spatial convolutional counterfactual regression (SC-CFR) shown in Figure \ref{fig:model}.
SC-CFR takes two inputs: covariates $\x$ and a treatment $z$, and outputs the outcome estimate $y$. 
SC-CFR first extracts effective spatial features $\Phi_{\x}$ from covariates $\x$ with the spatial feature extractor $g$.
Then, a multi-layer perceptron $h$ makes a prediction $y = h(\Phi_{\x}, \z)=h(g(\x), \z)$.

In the training phase of SC-CFR, to reduce the biases made by past decision-makers, we utilize the IPM regularizer that makes the distributions of the extracted representations $\Phi_{\x}=g(x)$ as similar as possible between different treatments.

\subsection{Spatial Feature Extraction}
Now, we explain the details of learning spatial features $\Phi_{\x}$.
We utilize spatial information, such as GPS locations of crowds for extracting spatial features.
We split spatial attributes vertically and horizontally to make a spatial discrete grid.
Then, from covariates $\x$, we construct a two-dimensional array $\ba \in \mathbb{R}^{d_1 \times d_2}$ filled with values such as the aggregated numbers of people in each block.

We employ a spatial convolution operator and an average pooling operator to extract spatial features $\Phi_{\x}$ from the two-dimensional array $\ba$.
These operators have been widely adopted in spatial data analysis, and have achieved significant improvements in various applications such as traffic forecasting \cite{ma2017learning,10.5555/3298239.3298479,Liang2019}.
These operators enable us to extract spatial distributions of the crowds by aggregating information of nearby grids, and the hierarchical structure of the operators captures both local and global interactions of the crowds.
Our two-dimensional spatial convolution operation \cite{goodfellow2016deep} takes the inner product or measures the cross-correlation between an input array $\ba$ and a kernel $\bk \in \mathbb{R}^{k_1 \times k_2}$, defined as
\begin{eqnarray}
 (\bk * \ba)(i, j) = \sum_{m=1}^{k_1} \sum_{n=1}^{k_2} \ba_{i + m, j + n} \bk_{m, n}.
\end{eqnarray}
The average pooling operator is the convolution operator whose kernel values are set to $1$.
We employ two repetitions of a convolution operator and an average pooling operator as $g(\x)$ in this paper (See Figure \ref{fig:model}).

\subsection{Balanced Representation}
If we do nothing to address the observation biases in data, the distribution of the extracted spatial feature $\Phi_{\x}$ is biased depending on $\z$, which could made the prediction accuracy of the model deteriorate.
To mitigate the bias and make $\Phi_{\x}$ a balanced representation, we utilize the integral probability metrics (IPM) \cite{muller1997integral,sriperumbudur2012empirical} in this paper.
Let us consider a binary treatment case $\mathcal{Z}=\{0, 1\}$: IPM can measure distance between two distributions.
Given two probability density functions $P(\Phi_{\x}|\z_i)$ and $P(\Phi_{\x}|\z_j)$, for $\mathcal{S} \subset \mathbb{R}^d$ and $G$, a family of functions from $ \mathcal{S}$ to $\mathbb{R}$, IPM is defined as
\begin{eqnarray}
{\rm IPM}_G(p, q) = \sup_{g \in G} \left|\int_{\mathcal{S}} g(s)\left(p(s) - q(s) \right) {\rm d}s \right|.
\end{eqnarray}
Intuitively, minimizing IPM makes the distributions of the extracted spatial features for the factual and counterfactual treatments hard to distinguish, and thus the distribution of the training data becomes close to those of RCTs \cite{shalit17a,ijcai2019-815}.

In this paper, we consider a case where a guiding plan $\z \in \mathcal{Z} =\{0,1\}^t $ consists of multiple actions, such as route recommendations for crowds and a modification of the road width.
However, the IPM compares only two distributions, and therefore, is not directly applicable to multiple treatments \cite{yoon2018ganite}.
One of the most natural extensions of IPM to multiple treatments is to measure IPM between all possible pairs of treatments; however, this requires  $|\mathcal{Z}|^2$ IPM terms and is not tractable for a large number of treatments.
To cope with this problem, we utilize another simple extension of IPM that considers the most frequent treatment in a mini-batch as the control treatment and only use the IPMs between the control and the others.
\begin{eqnarray}
  \sum_{\z' \in \mathcal{Z} \setminus \z}{\rm IPM}_G(p(\Phi_x|\z), q(\Phi_x|\z')).
\end{eqnarray}

\subsection{Outcome Prediction}
Finally, we utilize the spatial representations balanced for all the treatments to construct a predictive model $h(\Phi_{\x}, \z)$ to infer an outcome $\hat{y}$. 
The original counterfactual regression \cite{shalit17a} and its variants \cite{schwab2018perfect} employ multi-head networks where each multi-layer network corresponds to a treatment $z_i$ and shares common base layers to avoid the risk of losing treatment information. 
However, since the total number of treatments $|\mathcal{Z}|$ rapidly grows with the number of actions, and training all of the heads requires a considerable amount of observations that are not available in our setting, we employ a single-head network instead built on a multi-layer perceptron (MLP) that takes a concatenation of $\Phi_{\x}$ and $\z$ as the input.
\begin{eqnarray}
    h(\Phi_{\x}, \z) = {\rm MLP}(\Phi_{\x}, \z).
\end{eqnarray}
The single-head network architecture has showed better performance when the number of treatments is more than a dozen \cite{2006.05616}.

To train the SC-CFR model, we find the optimal parameter that minimizes the loss function.
\begin{eqnarray}
  \sum_{i=1}^N \| y_n -  {\rm MLP}(\Phi_{\x_n}, \z_n) \|_2^2 + \lambda \sum_{\z' \in \mathcal{Z}\setminus \z}{\rm IPM}_G(p(\Phi_x|\z_n), q(\Phi_x|\z'_n)),
\end{eqnarray}
where $\lambda >0$ is a hyper parameter for controlling the effect of IPM.
We employ the ADAM optimiser to optimize parameters in a mini-batch manner \cite{1412.6980}.
Among several candidates for the IPM, we employ the maximum mean discrepancy (MMD) \cite{JMLR:v13:gretton12a, sriperumbudur2012empirical} for its simplicity in this paper.

\begin{figure*}[tb]
  \begin{minipage}{4.cm}
  \centering
  \includegraphics[height=3cm]{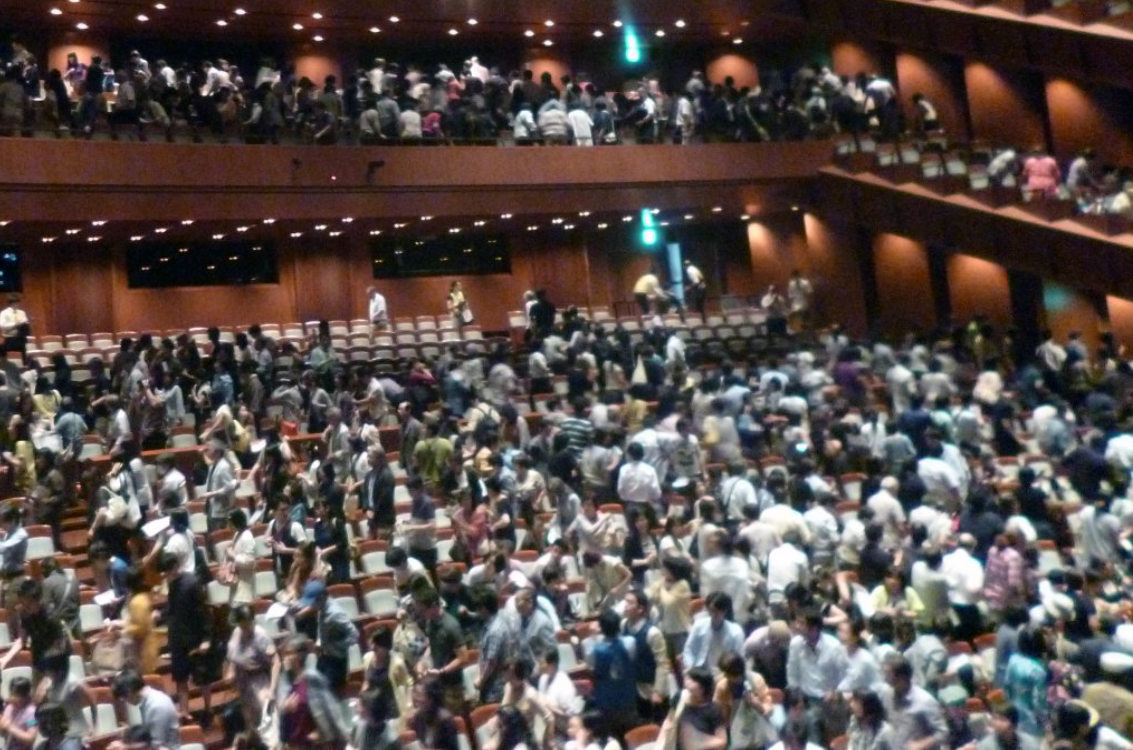}
  \caption{An evacuation.}
  \label{fig:drill-start}
  \Description{drill-start}
  \end{minipage}
  \quad\quad\quad\
  \begin{minipage}{4.5cm}
  \centering
  \includegraphics[height=3cm]{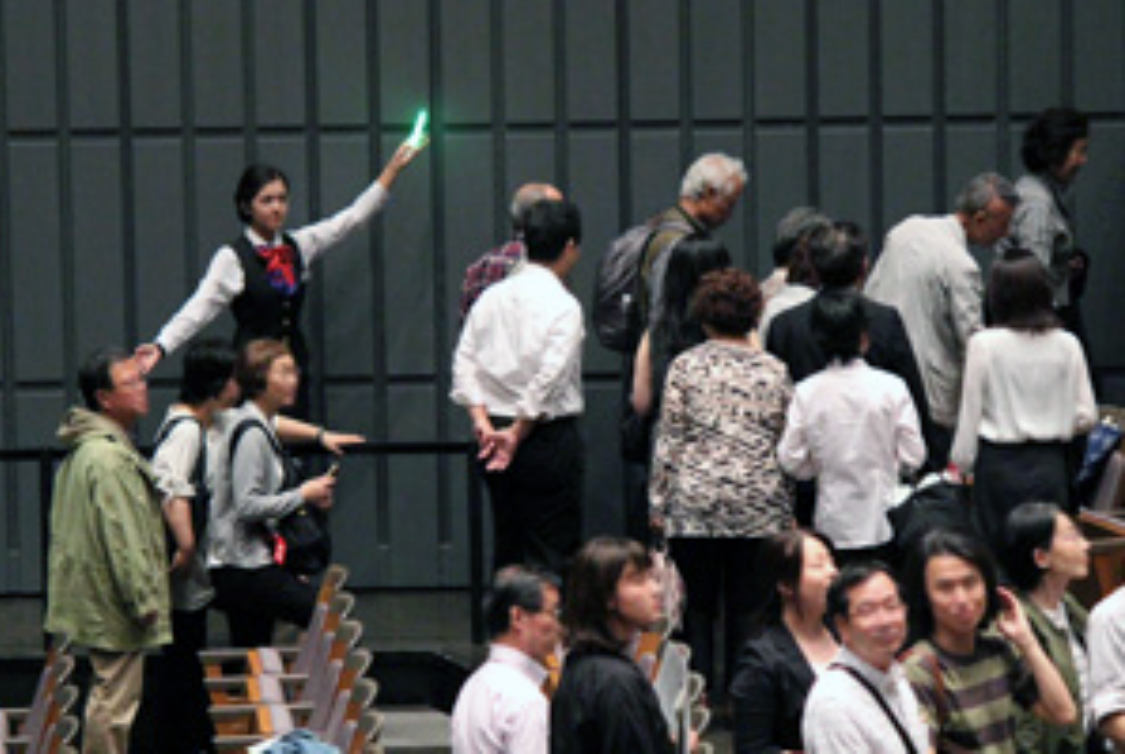}
  \caption{Route guidance.}
  \label{fig:drill-guide}
  \Description{root guide}
  \end{minipage}    
  \quad\quad\quad
  \begin{minipage}{4.5cm}
  \centering
  \includegraphics[height=3cm]{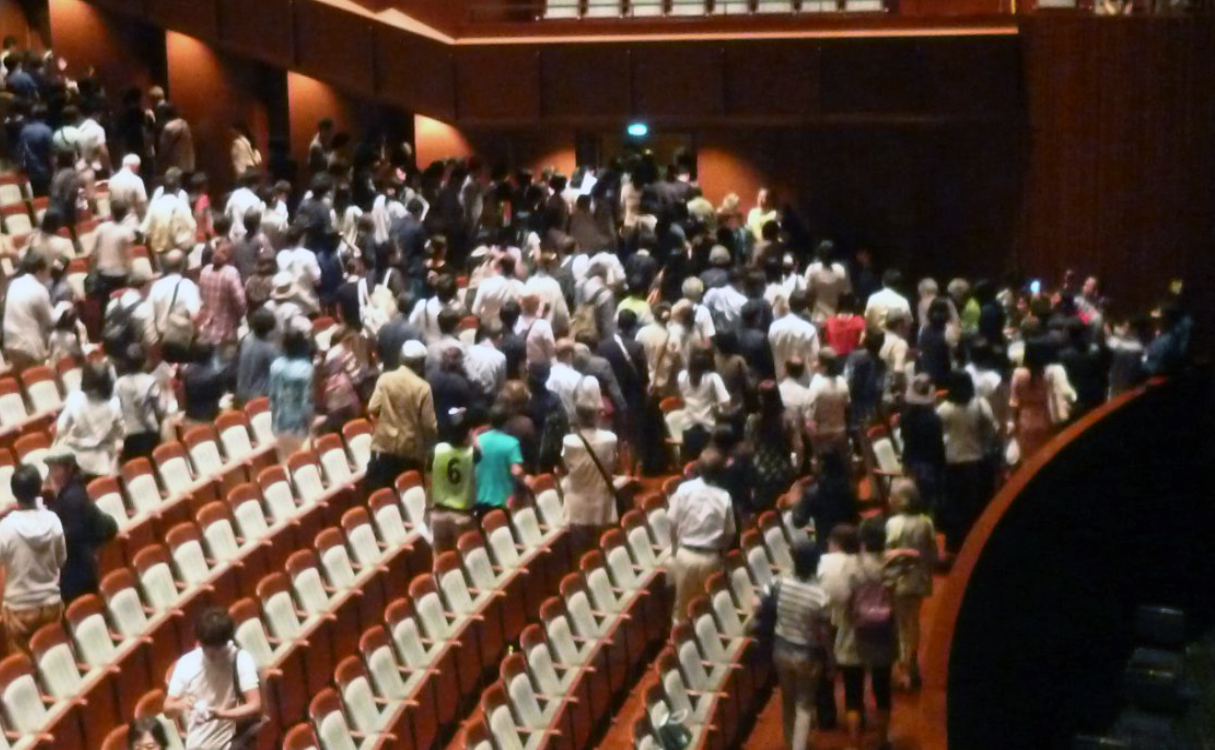}
  \caption{A jam of crowds.}
  \label{fig:drill-jam}
  \Description{Congestions}
  \end{minipage}
\end{figure*}

\section{Related Work}
Modeling and simulating large-scale crowd movements \cite{1373463}, including traffic network flows in cities \cite{varakantham2015direct,sharon2017real} and at sea \cite{singh2019multiagent}, and evacuation from buildings \cite{lammel2008bottlenecks,gianni2008simulation,tsai2011escapes,8823571} have been actively studied in the field of multi-agent systems.
Researchers have exploited multi-agent simulators to find optimal decision-making policies for crowd management.
Comparisons of the properties of various types of simulators are found in \cite{8365805}.
Most of the existing research focuses on bottom-up decision-making by individual crowds, i.e., each agent making decisions based on individual or partial information to achieve one's goals.
This study focuses on the indirect control of crowd movement through top-down decision-making by managers and the estimation of its effects.

A convolutional operator is a feature extraction method developed in computer vision for capturing low and high level image features \cite{goodfellow2016deep}.
Researchers in spatial data analysis have recently applied convolutional operators to handle the spatial correlations in urban areas \cite{ma2017learning}.
They have extracted local and city-wide global features by stacking spatial convolutional operators and utilized it to solve their domain-specific problems \cite{Liang2019}.
With such rich features, many researchers have reported significant improvements in the predictive accuracy on forecasting future amounts and speeds of transportation networks \cite{10.5555/3298239.3298479, Yu_2017} and rides of on-demand ride services \cite{KE2017591}, to name a few.
Based on their results, we utilize a spatial convolutional operator for extracting effective local spatial features of crowds from a small amount of data.

In the field of causal inference, novel methods for estimating the causal effect on individual and population levels have been proposed \cite{imbens2015causal,yao2020survey}.
The bayesian additive regression tree (BART) \cite{bart2011} is one of the most popular methods built on a simple decision tree model \cite{wager2018estimation, athey2019generalized} and it shows state-of-the-art performance.
Recently, in the field of machine learning,  novel methods based on deep neural networks to remove biases in data introduced by past decision-makers have been proposed \cite{10.5555/3045390.3045708, RePEc:ecl:stabus:3350,shalit17a,yao2018representation,Bica2020Estimating}; they use representation learning techniques to learn balanced representations of confounders.
In particular, the counterfactual regression (CFR) \cite{shalit17a} and its various extensions \cite{yoon2018ganite, schwab2018perfect,ijcai2019-815} have received widespread interest due to its significant performance improvement in estimating treatment effects compared to existing methods, such as BART.
However, most of the existing studies have not considered the use of spatial information to address data deficiencies and observation biases. 
In this paper, we use spatial convolution to extract effective spatial features for estimating treatment effects from the distribution of crowds given as images  \cite{ma2017learning,10.5555/3298239.3298479}.
The problem of estimating treatment effects is closely related to bandit feedback  \cite{strehl2010learning, beygelzimer2011contextual} and off-policy reinforce learning \cite{sutton2018reinforcement}.
Some of them assume that the policies of the agents used to collect data are known  \cite{imbens2015causal,ijcai2019-815}.
Meanwhile, we do not assume prior information about the data collection mechanisms.





\begin{figure}[tb]
  \centering
  \includegraphics[height=4.5cm]{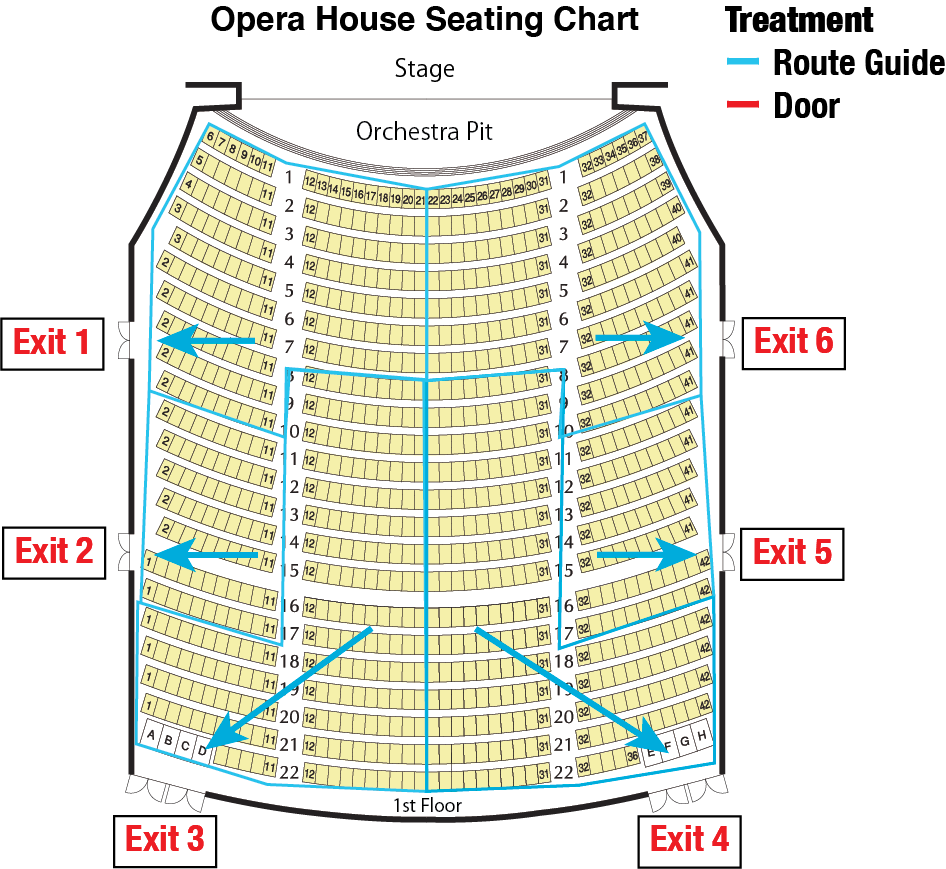}
  \centering
  \includegraphics[height=4.5cm]{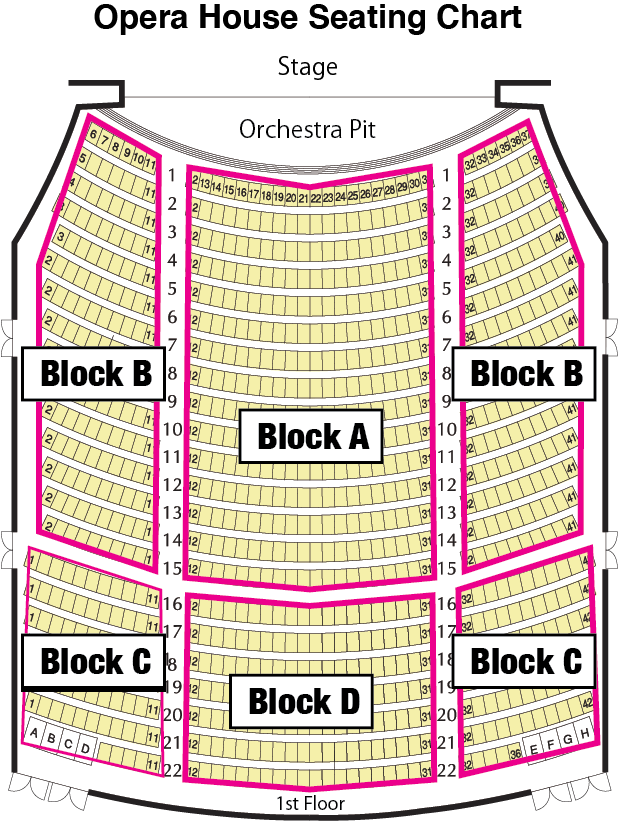}
  \caption{Building structures, seat layouts, and actions; a route guide for people (blue) and doors at six exits (red). Four blocks A, B, C, and D for data generation.}
  \label{fig:shinkoku}
  \Description{An example of evacuation routes.}
\end{figure}

\begin{figure*}[tb]
  \centering
  \includegraphics[width=16cm, trim=10mm 10mm 10mm 0mm]{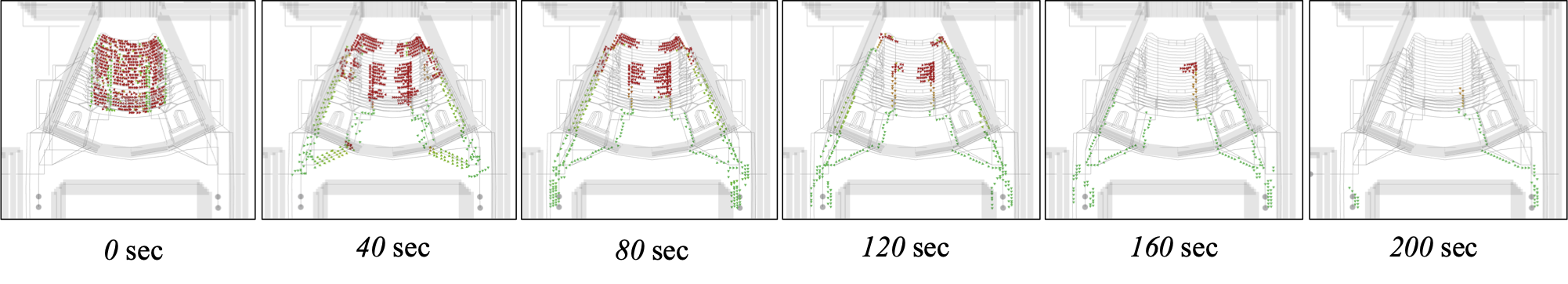}
  \caption{Snapshots of a multi-agent simulation without route guidance.}
  \label{fig:no_guide_effect}
  \Description{b}
\end{figure*}

\begin{figure*}[tb]
  \centering
  \includegraphics[width=16cm, trim=10mm 10mm 10mm 0mm]{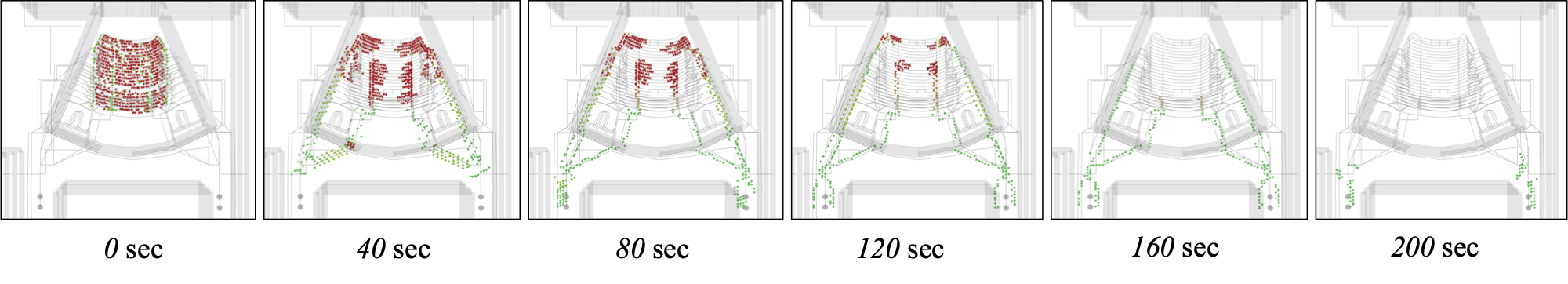}
  \caption{Snapshots of a multi-agent simulation with route guidance.}
  \label{fig:guide_effect}
  \Description{b}
\end{figure*}

\section{Experiments}
\subsection{Experimental Setup}
Evaluating the performance of estimating possible guidance treatment effects requires datasets recording the outcomes of all possible crowd movement guidance. 
Since such datasets are not available, we used an open-source multi-agent simulator \cite{Yamashita2012CrowdWalk} to generate realistic data on evacuation situations in a crowded theater.
We utilized the actual structure of an opera house, namely New National Theatre, Tokyo in our simulator, whose seating capacity is $d=868$.
We show the building structures in Figure \ref{fig:shinkoku}, in which the yellow blocks correspond to the seats. The blue lines and red texts represent actions of treatments explained, as in the following paragraph.
Crowds evacuate from the auditorium to the outside using one of the six exits. 
We conducted a real evacuation drill at the opera house and measured the evacuation movements of $570$ people using eight stereo cameras. 
Then, we employed the simulator \cite{Yamashita2012CrowdWalk} to generate crowd movement. 
We used the default setting parameters and set agents at the same initial locations as in the drill. 
We confirmed that the absolute error of the maximum evacuation time between measured and simulated data was less than five percent. 
We show images of an evacuation drill held in the opera house in Figures \ref{fig:drill-start}, \ref{fig:drill-guide}, and \ref{fig:drill-jam}.

In our experiments, we considered two types of actions.
The first type of action was {\em route guidance} to guide the audience to the evacuation exits.
If no route guide is placed, each audience evacuates from the nearest exit.
We employed a route guide whose maximum evacuation time was the smallest than other guides when all seats were occupied and all doors were open. 
The second type of the actions was {\em door operations} that determine whether the door of an exit to be fully open or half-open. 
In our scenario, we considered a risk, such as spreading fires, and allowed only two of six doors to be fully open.
We show snapshots of simulations having the same covariates but with and without route guidance in Figures \ref{fig:guide_effect} and \ref{fig:no_guide_effect}, where the green and red dots represent agents that are moving and stopping, respectively.
We can observe the positive effect of route guidance by comparing the snapshots at 160 sec; without route guidance, congestion occurred (Figure \ref{fig:no_guide_effect}), while it is dissolved if we give route guidance (Figure \ref{fig:guide_effect}).
Our codes and datasets are publicly available at a repository \footnote{https://github.com/koh-t/SC-CFR}.

\subsection{Data Generations}
\begin{figure*}[tb]
  \begin{minipage}{4cm}
  \centering
  \includegraphics[height=4cm]{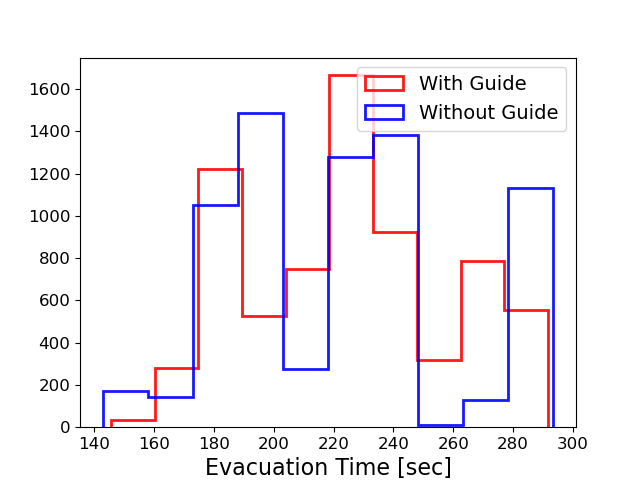}
  \caption{Maximum evacuation times.}
  \label{fig:max}
  \Description{Histogram of max outcomes.}
  \end{minipage}
  \qquad\qquad\quad
  \begin{minipage}{4cm}
  \centering
  \includegraphics[height=4cm]{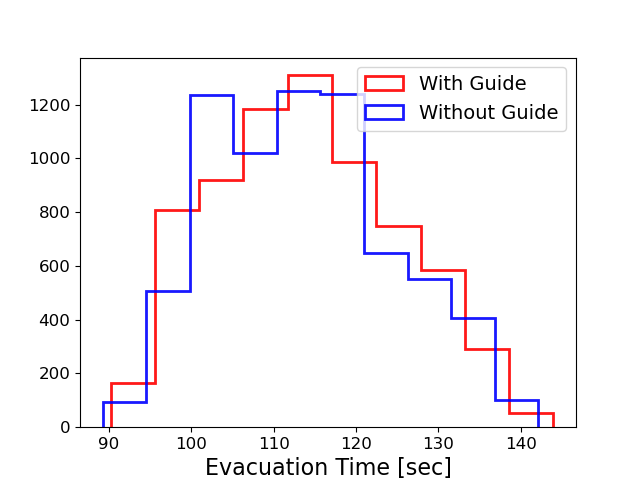}
  \caption{Average evacuation times.}
  \label{fig:ave}
  \Description{Histogram of average outcomes.}
  \end{minipage}    
  \qquad\qquad\quad
  \begin{minipage}{4cm}
  \centering
  \includegraphics[height=4cm]{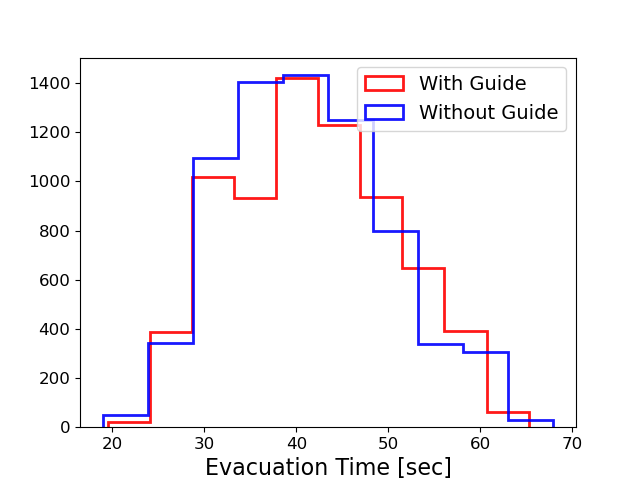}
  \caption{Standard deviation of the evacuation times.}
  \label{fig:std}
  \Description{Histogram of standard deviations outcomes.}
  \end{minipage}
\end{figure*}

We generated crowd movement datasets using our simulator with the following settings.
We first split the auditorium into four blocks (Blocks A, B, C, and D in Figure \ref{fig:shinkoku}) and set the occupancy rate of each block from $\{0.1, 0.5, 0.9\}$.
We randomly set the occupation of each seat with the occupancy rate and constructed covariates $\x$ by setting $x_i$ to $1$ if the $i$-th seat was occupied and set to $0$ otherwise.
We ran the seat selection ten times with different random seeds.
Then, we put agents on the occupied seats and ran evacuation simulations for all possible treatments $\z \in \mathcal{Z}$, where $|\mathcal{Z}|=2*(6*5)/2=30$.
We omitted simulations where the total number of people was less than $400$, since the guidance was not effective in such situations.
The total number of data $N$ was $423$ and each data has $30$ possible treatments and their corresponding outcomes.
We observed three outcomes for each run: the maximum, average, and standard deviation of evacuation times required by each person.
Then, we added observation noises to the outcomes that were randomly sampled from a normal distribution $\mathcal{N}(0, \sigma^2)$ where $\sigma=2$. 
We show the distributions of the outcomes in Figures \ref{fig:max}, \ref{fig:ave}, and \ref{fig:std}. 
The red and blue lines correspond to the outcomes with and without route guidance, respectively.

To select a factual treatment $\z$ for guidance, we employed two independent probabilistic policies (the propensity scores).
We set the first dimension of treatment $\z \in \{0,1\}^7$ to represent the action of the route guide, and others to indicate the actions of six doors.
We randomly determined whether to use the route guide using a distribution $p(z_{\rm guide}=1) = 1/(1 + \exp(-\sum_{i=1}^d x_i / d + 1))$.
We set $z_{\rm guide}$ to $1$ when the route guidance is given.
To decide which exit doors to be fully open, we calculated the populations of occupied seats within $8$ meters $\sum_{i \in \mathcal{D}_{j}}x_i/|\mathcal{D}_{j}|$ from each door, where $\mathcal{D}_j$ is the number of the seats within $8$ meters from the $j$-th door.
Finally, we randomly sampled two doors to be open by using a multinomial distribution whose parameters were set to those populations.
We treated all the other treatments as counterfactual treatments.


\subsection{Model Settings}
\begin{figure}[tb]
  \centering
  \includegraphics[width=3cm]{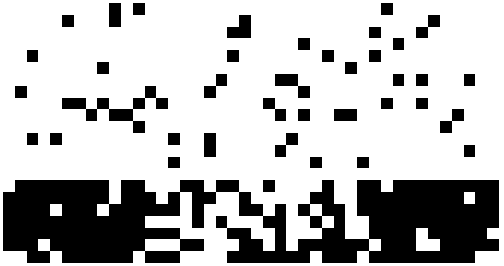}
  \qquad
  \includegraphics[width=3cm]{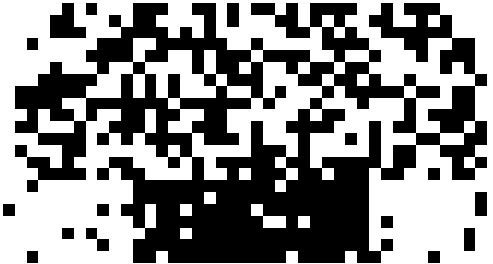}\\
  \vspace*{0.5cm}
  \includegraphics[width=3cm]{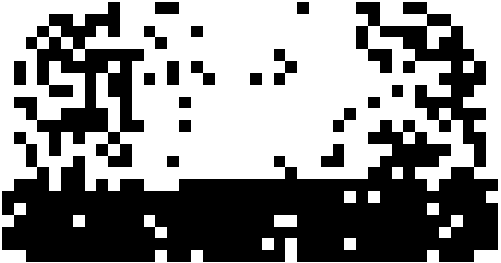}
  \qquad
  \includegraphics[width=3cm]{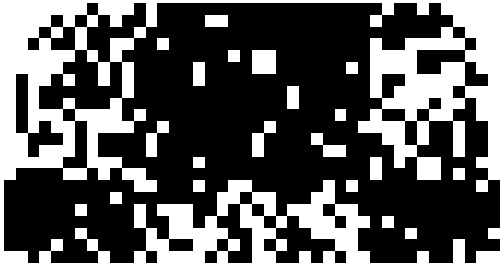}
  \caption{Examples of two-dimensional covariate arrays.}
  \label{fig:twodimarray}
  \Description{Example of covariate arrays.}
\end{figure}

We randomly sampled $90\%$ of the seat assignments as a training dataset and used the rest as a test dataset.
As the baseline methods, we employed nine supervised learning methods \cite{friedman2001elements}: Lasso \cite{tibshirani1996regression}, the ridge regression (Ridge), the support vector regression (SVR), the random forest (RF), the gradient boosting regressor (GBR), the bayesian additive regression tree (BART) \cite{bart2011}, and the multi-layer perceptron (MLP) consisting of five layers. 
Since these methods were not applicable to identify differences between covariates and a treatment, we simply utilized a concatenation $[\x, \z]$ as an input.
The treatment-agnostic representation network (TARNET) and the counterfactual regression (CFR) are recently proposed neural network-based causal effect estimation methods in machine learning fields \cite{shalit17a}.
For TARNET and CFR, we used a two-layer MLP to learn the representation of $\x$ and a three-layer MLP whose inputs were a concatenation of the extracted representation and treatments to predict the outcome.
We employed two variants of our proposed methods.
SC-CFR is our proposed method using IPM, and a spatial convolutional treatment-agnostic representation network (SC-TARNET) is the proposed method without IPM ($\lambda=0)$. 
We constructed two-dimensional arrays $\bm{a} \in \mathbb{R}^{22 \times 42}$ from covariates based on seat layouts. We show examples of the arrays in Figure \ref{fig:twodimarray}.
For our methods, we utilized two repetitions of a block that consisted of a convolutional operator and an average pooling operator, whose kernel size $k$ and paddings were set to $3$ to extract the spatial feature $\Phi_{\x}$.
Then, we used a three-layer MLP to predict the outcome.
We selected model architectures and hyperparameters by minimizing the training errors.

\subsection{Metrics}

To compare the performance of models, we employed several metrics that could measure the accuracy of predicted outcomes $\hat{y}$ under the potential outcome framework. 
We evaluated models in two different settings; 
the {\em within-sample} setting, where the task is to estimate outcomes for all scenarios in the training dataset, evaluates what happens if a decision-maker chose other treatments in the past evacuation events.
On the other hand, the {\em out-of-sample} setting, where the task is to estimate outcomes for scenarios in the test dataset, corresponds to the case of a new evacuation event, and the goal is to estimate the outcomes of all possible treatments.

{\bf Root mean squared error (RMSE)}.
RMSE is a widely used metric in supervised learning problems. 
We employed RMSE to measure the accuracy of each predicted outcome depending on the covariates and all possible treatments. 
We define $\epsilon_{\rm RMSE}$ as
\begin{eqnarray}
    \left[\frac{1}{|\mathcal{Z}|}\frac{1}{N} \sum_{\z_i \in \mathcal{Z}} \sum_{n=1}^N \left( \hat{y}_{\x_n}(\z_i) -  \mathbb{E}[Y_{\x_n}(\z_i)]\right)^{2}\right]^{1/2},
\end{eqnarray}
where $\mathcal{Z}'$ is a set of whole pairs of possible guiding plans.

{\bf Precision in estimation of heterogeneous effect (PEHE)}.
In the binary setting, PEHE measures the ability of a predictive model to estimate the difference between true and predicted outcomes between two treatments $z_0$ and $z_1$ for covariates $x$. 
We define PEHE as the RMSE between the true difference of the outcomes of two treatments $\z_i$ and $\z_j$, that is, $\mathbb{E}[Y_{\x_n}(\z_i)] - \mathbb{E}[Y_{\x_n}(\z_j)]$, and the predicted difference, $\hat{y}_{\x_n}(\z_i) - \hat{y}_{\x_n}(\z_j)$.
Obtaining a better PEHE requires accurate estimations of both the factual and counterfactual outcomes. 
We set $\epsilon_{\rm PEHE}$ as
\begin{eqnarray}
    \left[ \frac{1}{N}\sum_{n=1}^N \left( \hat{y}_{\x_n}(\z_i) - \hat{y}_{\x_n}(\z_j) - \left(\mathbb{E}[Y_{\x_n}(\z_i)] 
    - \mathbb{E}[Y_{\x_n}(\z_j)]\right)\right)^{2}\right]^{1/2}.
\end{eqnarray}

{\bf Average treatment Effect (ATE)}.
We employed the error on estimating a conditional ATE. 
In contrast to PEHE, ATE considers only the average difference of treatment effects over the population. 
ATE is not as important as PEHE for models optimised for CATE estimation, but it can be a useful indicator of how well an estimator performs at comparing two treatments across the entire population. 
We define $\epsilon_{\rm ATE}$ as
\begin{eqnarray}
    \frac{1}{N}\sum_{n=1}^N \left( \hat{y}_{\x_n}(\z_i) - \hat{y}_{\x_n}(\z_j)\right)
    - \frac{1}{N}\sum_{n=1}^N\left(\mathbb{E}[Y_{\x_n}(\z_i)] - \mathbb{E}[Y_{\x_n}(\z_j)]\right).
\end{eqnarray}

{\bf Multiple Treatments}.
Let us consider a general case where combinations of multiple actions are available for making a guiding plan.
By following \cite{schwab2018perfect}, we extended PEHE and ATE to be applicable to multiple treatments by considering the average PEHE and ATE between every possible pair of guiding plans defined as
\begin{eqnarray}
 \epsilon_{\rm mPEHE} = \frac{1}{\binom{t}{2}}\sum_{(\z_i, \z_j) \in \mathcal{Z}} \epsilon_{\rm PEHE},~~
 \epsilon_{\rm mATE} = \frac{1}{\binom{t}{2}}\sum_{(\z_i, \z_j) \in \mathcal{Z}} \epsilon_{\rm ATE}.
\end{eqnarray}

\begin{table*}[tb]
  \caption{Results on estimating the maximum evacuation time.}
  \label{tab:max}
  \small
\centering
{\tabcolsep=0.7em
\begin{tabular}{@{\extracolsep{4pt}}lrrrrrr}
\toprule
\multirow{2}{*}{Method} &  \multicolumn{3}{c}{Within-sample} &  \multicolumn{3}{c}{Out-of-sample} \\
\cline{2-4}\cline{5-7}
{} & $\epsilon_{\rm RMSE}$ & $\epsilon_{\rm mPEHE}$ & $\epsilon_{\rm mATE}$  
& $\epsilon_{\rm RMSE}$ & $\epsilon_{\rm mPEHE}$ & $\epsilon_{\rm mATE}$\\
\midrule
Lasso & $17.334_{(0.282)}$  & $19.047_{(0.452)}$  & $5.604_{(0.327)}$ & $18.524_{(1.216)}$ & $19.290_{(0.777)}$ & $5.947_{(0.674)}$\\
Ridge & $16.886_{(0.576)}$  & $18.396_{(0.317)}$  & $7.216_{(0.642)}$ & $24.369_{(1.962)}$ & $18.575_{(0.841)}$ & $7.276_{(0.841)}$\\
SVR & $14.895_{(0.179)}$ & $19.125_{(0.159)}$  & $5.534_{(0.138)}$ & $15.519_{(0.714)}$ & $19.365_{(0.892)}$ & $5.860_{(0.469)}$\\
RF & $13.595_{(0.444)}$ & $16.117_{(0.613)}$ & $4.952_{(0.362)}$ & $14.824_{(0.779)}$ & $16.406_{(0.906)}$ & $5.234_{(0.667)}$\\
GBR & $11.617_{(0.503)}$ & $13.933_{(0.577)}$  & $3.492_{(0.675)}$ & $12.841_{(0.890)}$  & $14.290_{(1.221)}$ & $3.966_{(1.079)}$\\
BART & $11.620_{(1.299)}$ & $14.429_{(1.601)}$  & $4.541_{(1.324)}$ & $12.508_{(1.375)}$  & $14.796_{(1.840)}$  & $4.863_{(1.559)}$\\
MLP  & $17.911_{(0.310)}$ & $19.125_{(0.286)}$  & $5.381_{(0.235)}$ & $21.243_{(1.628)}$ & $19.342_{(0.781)}$   & $5.656_{(0.511)}$\\
\midrule
TARNET  & $12.992_{(0.227)}$ & $15.115_{(0.281)}$  & $3.782_{(0.576)}$  & $13.981_{(0.863)}$ & $15.433_{(0.797)}$  & $4.147_{(0.923)}$\\
CFR  & ${}^{**}12.660_{(0.576)}$ & ${}^{*}14.399_{(0.713)}$ & $3.658_{(0.701)}$ & ${}^{**}13.511_{(1.052)}$ & ${}^{*}14.716_{(1.259)}$  & $4.025_{(1.057)}$\\
\midrule
SC-TARNET & $7.748_{(0.746)}$ & $9.767_{(1.029)}$ & $3.017_{(0.506)}$ & $7.920_{(0.595)}$  & $9.932_{(0.938)}$  & $3.288_{(0.738)}$\\
SC-CFR & ${}^{**}\bm{7.658_{(0.787)}}$ & ${}^{*}\bm{9.636_{(1.032)}}$ & ${}^{*}\bm{2.933_{(0.512)}}$ & ${}^{**}\bm{7.816_{(0.579)}}$ & ${}^{*}\bm{9.802_{(0.902)}}$ & ${}^{**}\bm{3.209_{(0.761)}}$\\
\bottomrule
\end{tabular}
}

\end{table*}

\begin{table*}[tb]
  \caption{Results on estimating the average evacuation time.}
  \label{tab:mean}
  \small
\centering
{\tabcolsep=0.9em
\begin{tabular}{@{\extracolsep{4pt}}lrrrrrr}
\toprule
\multirow{2}{*}{Method} &  \multicolumn{3}{c}{Within-sample} &  \multicolumn{3}{c}{Out-of-sample} \\
\cline{2-4}\cline{5-7}
{} & $\epsilon_{\rm RMSE}$ & $\epsilon_{\rm mPEHE}$ & $\epsilon_{\rm mATE}$  
& $\epsilon_{\rm RMSE}$ & $\epsilon_{\rm mPEHE}$ & $\epsilon_{\rm mATE}$\\
\midrule
Lasso & $6.497_{(0.167)}$ & $3.717_{(0.075)}$  & $2.133_{(0.106)}$  & $6.889_{(0.604)}$ & $3.743_{(0.155)}$  & $2.168_{(0.140)}$\\
Ridge & $3.075_{(0.096)}$ & $2.726_{(0.108)}$  & $0.758_{(0.236)}$ & $5.034_{(0.448)}$  & $2.708_{(0.176)}$  & $0.720_{(0.257)}$\\
SVR & $2.860_{(0.093)}$ & $2.747_{(0.074)}$ & $1.188_{(0.098)}$ & $2.859_{(0.166)}$ & $2.765_{(0.161)}$ & $1.185_{(0.165)}$\\
RF & $3.185_{(0.105)}$ & $3.293_{(0.148)}$  & $1.819_{(0.118)}$ & $3.627_{(0.203)}$ & $3.321_{(0.178)}$ & $1.848_{(0.162)}$\\
GBR & $2.336_{(0.145)}$ & $2.255_{(0.172)}$ & $0.810_{(0.163)}$ & $2.793_{(0.209)}$ & $2.277_{(0.161)}$ & $0.827_{(0.155)}$\\
BART & $2.298_{(0.133)}$ & $2.296_{(0.128)}$ & $\bm{0.507_{(0.076)}}$ & $2.503_{(0.215)}$ & $2.328_{(0.160)}$ & $\bm{0.523_{(0.074)}}$\\
MLP & $4.132_{(0.078)}$ & $3.323_{(0.098)}$ & $1.580_{(0.112)}$ & $5.688_{(0.543)}$ & $3.361_{(0.139)}$ & $1.611_{(0.171)}$\\
\midrule
TARNET & $2.589_{(0.077)}$ & $2.470_{(0.042)}$ & $0.761_{(0.159)}$ & $3.030_{(0.301)}$ & $2.494_{(0.145)}$  & $0.811_{(0.146)}$\\
CFR & ${}^{**}2.537_{(0.074)}$ & ${}^{**}2.337_{(0.065)}$ & ${}^{**}0.702_{(0.110)}$ & ${}^{**}3.014_{(0.329)}$  & $2.357_{(0.148)}$  & $0.731_{(0.115)}$\\
\midrule
SC-TARNET & $1.849_{(0.240)}$ & $1.947_{(0.221)}$ & $0.569_{(0.105)}$ & $1.929_{(0.252)}$  & $1.976_{(0.224)}$  & $0.602_{(0.106)}$\\
SC-CFR & $\bm{1.922_{(0.250)}}$  & $\bm{1.816_{(0.260)}}$  & $0.542_{(0.106)}$ & $\bm{1.869_{(0.237)}}$ & $\bm{1.955_{(0.208)}}$ & $0.572_{(0.085)}$\\
\bottomrule
\end{tabular}
}

\end{table*}

\begin{table*}[tb]
  \caption{Results on estimating the standard deviation of evacuation time.}
  \label{tab:std}
  \small
\centering
{\tabcolsep=0.9em
\begin{tabular}{@{\extracolsep{4pt}}lrrrrrr}
\toprule
\multirow{2}{*}{Method} &  \multicolumn{3}{c}{Within-sample} &  \multicolumn{3}{c}{Out-of-sample} \\
\cline{2-4}\cline{5-7}
{} & $\epsilon_{\rm RMSE}$ & $\epsilon_{\rm mPEHE}$ & $\epsilon_{\rm mATE}$  
& $\epsilon_{\rm RMSE}$ & $\epsilon_{\rm mPEHE}$ & $\epsilon_{\rm mATE}$\\
\midrule
Lasso & $6.538_{(0.130)}$  & $4.957_{(0.103)}$  & $2.861_{(0.108)}$  & $6.699_{(0.472)}$ & $5.012_{(0.294)}$ & $2.950_{(0.231)}$\\
Ridge & $3.571_{(0.109)}$  & $3.403_{(0.038)}$  & $0.852_{(0.099)}$  & $5.487_{(0.384)}$ & $3.426_{(0.198)}$  & $0.941_{(0.208)}$\\
SVR  & $3.455_{(0.100)}$ & $3.658_{(0.087)}$  & $1.687_{(0.099)}$  & $3.420_{(0.139)}$ & $3.694_{(0.253)}$   & $1.730_{(0.225)}$\\
RF & $3.412_{(0.237)}$  & $3.730_{(0.258)}$  & $1.885_{(0.231)}$ & $3.649_{(0.250)}$  & $3.756_{(0.197)}$  & $1.905_{(0.257)}$\\
GBR & $2.565_{(0.164)}$  & $2.693_{(0.147)}$  & $0.790_{(0.176)}$ & $2.832_{(0.183)}$  & $2.720_{(0.117)}$  & $0.898_{(0.161)}$\\
BART & $2.567_{(0.098)}$ & $2.732_{(0.180)}$  & $0.537_{(0.053)}$ & $2.772_{(0.292)}$ & $2.764_{(0.168)}$ & $0.658_{(0.101)}$\\
MLP & $4.427_{(0.127)}$  & $4.488_{(0.113)}$  & $2.353_{(0.114)}$ & $5.157_{(0.314)}$  & $4.517_{(0.220)}$  & $2.395_{(0.175)}$\\
\midrule
TARNET  & $2.961_{(0.144)}$ & $3.053_{(0.150)}$  & $0.786_{(0.116)}$ & $3.280_{(0.350)}$  & $3.087_{(0.200)}$  & $0.888_{(0.201)}$\\
CFR  & ${}^{**}2.866_{(0.124)}$ & ${}^{**}2.772_{(0.215)}$  & ${}^{*}0.734_{(0.092)}$ & ${}^{**}3.131_{(0.316)}$ & $2.834_{(0.201)}$ & $0.838_{(0.174)}$\\
\midrule
SC-TARNET & $1.889_{(0.159)}$  & $2.095_{(0.176)}$  & $0.561_{(0.123)}$ & $\bm{1.933_{(0.239)}}$ & $2.100_{(0.209)}$ & $0.652_{(0.170)}$\\
SC-CFR & $\bm{1.867_{(0.154)}}$ & $\bm{2.062_{(0.174)}}$ & ${}^{*}\bm{0.525_{(0.103)}}$ & $1.954_{(0.229)}$  & $\bm{2.081_{(0.221)}}$ & ${}^{**}\bm{0.609_{(0.138)}}$\\
\bottomrule
\end{tabular}
}

\end{table*}

\subsection{Results}
Tables \ref{tab:max}, \ref{tab:mean}, and \ref{tab:std} show the results of prediction of the maximum, average, and standard deviation of evacuation times,  respectively.
Each column represents the average and the standard deviation of each metric. 
A bold letter indicates a method that achieved the smallest (best) error in each column.
To assess the effect on bias cancellation, we employed the t-test on two related samples of the scores between TARNET and CFR, and SC-TARCNET and SC-CFR.
The values with a single star or double stars indicate that the p-values between the two models were $p<0.10$ and $p<0.05$, respectively.

From Tables \ref{tab:max}, \ref{tab:mean}, and \ref{tab:std}, we confirmed that our proposed SC-CFR achieved the best performance on $\epsilon_{\rm RMSE}$ and $\epsilon_{\rm mPEHE}$.
SC-TARNET came in the second place. 
GBR or BART placed third.
In Table \ref{tab:max}, SC-CFR reduced the estimation error of $\epsilon_{\rm mPEHE}$ out-of-sample by $33\%$ and $38\%$ from GBR, and $55\%$ and $56\%$ from Ridge, respectively.

The improvements of SC-CFR on $\epsilon_{\rm RMSE}$ and $\epsilon_{\rm mPEHE}$ for both within-sample and out-of-sample settings indicated that our method successfully extracted spatial features and learned balanced representations, thus providing more accurate predictions of outcomes than the  state-of-the-art baselines.
Our methods achieved the best estimation results of $\epsilon_{\rm mATE}$ in Tables \ref{tab:max} and \ref{tab:std}, while BART showed the best performance in Table \ref{tab:mean}.
Since the distribution of the standard deviations had a relatively simpler shape than the others, BART, which is a simpler non-linear method than ours, was sufficient to estimate ATE.



By comparing MLP and TARNET, TARNET showed better scores than MLP because of its representation learning mechanism, which MLP does not have. 
The performance of SC-CFR was significantly better than that of CFR. 
With these results, we could argue that the spatial convolutional operators worked correctly and greatly contributed to this improvement.
As regarding the effect of bias cancellation, we observed that IPM contributed to the performance improvements of CFR from TARNET and SC-CFR from SC-TARNET.
This result indicates that a balanced representation was effective in precisely estimating CATE of crowd movement guidance, which is consistent with existing literature \cite{shalit17a}.



\section{Conclusion}
In this paper, we focused on the problem of estimating the effects of crowd movement guidance from a relatively small amount of biased data to answer a typical question posed by decision-makers: {\em ``what will be the outcomes of each of the possible options in the current situation?"}.
We proposed a Spatial Convolutional Counterfactual Regression (SC-CFR), which takes advantage of two recent developments in deep representation learning in spatial data analysis and causal inference.
We utilized spatial convolutional operators to extract effective spatial features of crowds from a small amount of data, and applied balanced representation learning using IPM to mitigate the selection bias and missing counterfactual outcomes.
With datasets generated by a multi-agent simulator on evacuation scenarios in a crowded theater, we conducted experiments to estimate treatment effects on three types of outcomes of evacuation times required by crowds.
We demonstrated that SC-CFR reduced the estimation error by at most $56\%$ from the dstate-of-the-art baseline methods that did not consider spatial attributes.

{\bf Acknowledgement} This work was supported by JST, PRESTO Grant Number JPMJPR20C5, Japan and JSPS KAKENHI Grant Numbers 20H00609, Japan




\bibliographystyle{ACM-Reference-Format} 
\balance
\bibliography{simulator}


\end{document}